\documentclass[5p]{elsarticle}

\usepackage{lineno,hyperref}
\usepackage{subfig}
\modulolinenumbers[5]

\journal{Journal of \LaTeX\ Templates}

\biboptions{authoryear}

\usepackage{amsmath}
\usepackage{booktabs}
\usepackage{multirow}

\usepackage{subfig}
\usepackage{siunitx}
\usepackage{makecell}
\usepackage{amssymb}
\usepackage{tabularx}
\usepackage[ruled,linesnumbered]{algorithm2e}

\begin{document}

\begin{frontmatter}

	\title{GWRBoost:A geographically weighted gradient boosting method for explainable quantification of spatially-varying relationships}

	\author[add1,add2]{Han Wang}
    \author[add1,add2]{Zhou Huang\corref{mycorrespondingauthor}}

	\cortext[mycorrespondingauthor]{Corresponding author}
	\ead{huangzhou@pku.edu.cn}
    
    \author[add1,add2]{Ganmin Yin}
    \author[add1,add2]{Yi Bao}
    \author[add1,add2]{Xiao Zhou}
    \author[add1,add2]{Yong Gao}

	\address[add1]{Institute of Remote Sensing and Geographical Information Systems, Peking University, Beijing, China}
	\address[add2]{Beijing Key Lab of Spatial Information Integration \& Its Applications, Peking University, Beijing, China}

	\begin{abstract}
		The geographically weighted regression (GWR) is an essential tool for estimating the spatial variation of relationships between dependent and independent variables in geographical contexts. However, GWR suffers from the problem that classical linear regressions, which compose the GWR model, are more prone to be underfitting, especially for significant volume and complex nonlinear data, causing inferior comparative performance. Nevertheless, some advanced models, such as the decision tree and the support vector machine, can learn features from complex data more effectively while they cannot provide explainable quantification for the spatial variation of localized relationships. To address the above issues, we propose a geographically gradient boosting weighted regression model, GWRBoost, that applies the localized additive model and gradient boosting optimization method to alleviate underfitting problems and retains explainable quantification capability for spatially-varying relationships between geographically located variables. Furthermore, we formulate the computation method of the Akaike information score for the proposed model to conduct the comparative analysis with the classic GWR algorithm. Simulation experiments and the empirical case study are applied to prove the efficient performance and practical value of GWRBoost. The results show that our proposed model can reduce the RMSE by 18.3\% in parameter estimation accuracy and AICc by 67.3\% in the goodness of fit.
	\end{abstract}

\end{frontmatter}


\section{Introduction}
It is essential to consider spatial heterogeneity of relationships between located variables in the analysis of any geographical process, which is also referred to as spatial nonstationarity \citep{goodchild2004validity}. The quantification analysis of the relationship between the response variable and predictor variables varies among the observed data samples in the study area. Classic models are insufficient to evaluate the spatialized observations. Extensive methods have been proposed to investigate the spatial nonstationarity and improve the fitting performance of analytical models \citep{anselin1995local, haas1996multivariate, getis2010analysis}. The geographically weighted regression (GWR) is one of the localized models to evaluate the spatial variation in the relationships between the dependent and independent variables \citep{brunsdon1996geographically}. Instead of assuming constant geographical processes over space, the GWR model investigates the spatial heterogeneity of correlations by applying a local linear model for each observation and the weighted least square algorithm to estimate coefficients to represent the relationship between the dependent and independent variables \citep{huang2010stgwr}. Numerous spatialized studies leverage GWR to explore the characteristics of spatial non-stationary, such as environmental research \citep{zhou2022identifying}, transportation analysis \citep{xu2019geographically, ZHOU2022104348}, socio-economic studies \citep{wang2019analyzing}.

In the analysis of spatialized explanatory variables, GWR provides advantages over the global model in various aspects. Generally, GWR has a better fitting and predictive capabilities for sampled and unsampled locations \citep{kupfer2007incorporating}. The internal characteristics in complicated spatial data can be captured by local estimates. And the group of local models with a higher model complexity can achieve a better comparative performance than global linear regression. Furthermore, local spatial non-stationary explanatory variables relationships, which may be hidden by the global model, can be detected by localized methods, including GWR \citep{fotheringham2003gwr}. The spatial variation in the relationships between variables provides a new perspective to investigate the spatial heterogeneity of effects on the response variable. GWR can also minimize spatial autocorrelation of residuals, which indicates the insufficient capability of a global model due to non-stationary \citep{zhang2005spatial}.

Although GWR can be utilized to analyze spatially varying relationships between variables, some limitations still need to be addressed and corresponding modifications have been proposed to mitigate these disadvantages. First, fewer samples in the neighbouring region of the location to be estimated will incur a higher risk of multicollinearity \citep{wheeler2005multicollinearity}. Geographically weighted ridge regression \citep{wheeler2007diagnostic} and geographically weighted lasso \citep{wheeler2009simultaneous} alleviate the multicollinearity issue by applying ridge and lasso regression respectively. Additionally, the spatial variation of parameters is the primary limitation for testing hypotheses, which can mislead the analysis of regional relationships \citep{jetz2004local}. The reused local samples in a different location also contribute to the multiple testing issue \citep{da2016multiple}.

Despite a variety of modifications and improvements to mitigate the addressed problems, little is still known about several issues of GWR. First, the linear model at each sampled location aims to optimize the local weighted squared error independently while most evaluation criteria focus on measuring the global squared error, including root mean squared error (RMSE), r-squared (R$^2$) and Akaike information criterion score (AIC) \citep{akaike1998aic}. Local models roughly maintain the same optimization direction as the global model in the objective function parameter space but do not necessarily obtain the optimal solutions. Second, traditional linear regression is prone to be underfitting, especially for big volume and complex nonlinear data, causing inferior comparative performance \citep{michael1997lrunderfit}. Several studies use more complex machine learning algorithms to enhance the fitting performance, but they are not capable of quantifying the spatially varying relationship between variables explicitly, like the estimation of the coefficients in the linear model \citep{xu2019svm}. Additionally, the AIC score, which is widely used in the geographically fitting procedure, cannot be applied to evaluate the model complexity of nonparametric models, such as decision trees and support vector machines (SVM). Models with higher complexity may suffer from undetected overfitting problems due to the absence of proper evaluation metrics, such as AIC and AICc, or the train-test split training paradigm. Thus, more complex and non-differentiable models have a higher risk of overfitting and are not necessarily better than traditional GWR if evaluated by the metric of AIC.

To address these issues, we propose \textbf{GWRBoost}, a geographically weighted regression method that applies the localized additive model and gradient boosting optimization method to alleviate underfitting problems. The GWRBoost leverages classic squared error as the objective function, which is compatible with the evaluation criteria. In the optimization process, we use gradient boosting as a stage-wise method to obtain the global optimal parameters, which employs geographically weighted linear regression as the base procedure. Furthermore, the GWRBoost maintains a linear regression form to retain the ability of explainable quantification for spatially-varying relationships between geographically located variables. And it is possible to evaluate the performance of GWRBoost by the AIC/AICc score, which can be compared with traditional GWR.

This study makes three main contributions. First, we leverage the additive model for located observations and propose the GWRBoost algorithm. Second, we formulate the computation of the degree of freedom for the GWRBoost model. Third, simulation experiments and an empirical case study are applied to prove the performance and practical value of GWRBoost. The rest of this paper is organized as follows. Section~\ref{sec:review} reviews existing research on variants of geographically weighted regression and gradient boosting models. Section~\ref{sec:method} presents the GWRBoost algorithm. Section~\ref{sec:expr} introduces the experiments on simulated data and the case study. The discussion is given in Section~\ref{sec:discuss}. And the conclusions are presented in Section~\ref{sec:conclusion}.

\section{Related work}
\label{sec:review}
\subsection{Variants of geographically weighted regression}
Numerous variants have been developed to improve the GWR in various aspects. Several studies focus on the improvement of optimal bandwidth selection. Generally, the choice of bandwidth is crucial to the fitting result of GWR \citep{fotheringham2003gwr}. A larger bandwidth smooths the variation of local parameter estimates but invites larger bias, whereas a smaller bandwidth will make them sharper with larger variance \citep{lu2018improvements}. An optimal bandwidth can be acquired by searching with common evaluation metrics, including cross-validation and AIC/AICc.

To relax the scale assumptions for bandwidth of different explanatory variables, \cite{brunsdon1999sgwr} designed a semiparametric GWR (SGWR), which combines global relationships without distance decay effect and local relationships estimated by weighted regression. \cite{fotheringham2017mgwr} further proposed a multiscale geographically weighted regression (MGWR), which assumes that dependent variables have different scales of spatial heterogeneity. Thus, the MGWR model, which is more flexible than SGWR, applies individual bandwidth for each dependent variable and is optimized by a component-wise back-fitting algorithm. Some empirical studies showed that the MGWR performs better than GWR on real world data sets \citep{zhou2022identifying}.

Some researchers concentrate on the modification of spatial distance metrics. \cite{lu2014geographically} first applied non-Euclidean distance metrics for GWR models, including road network distance and travel time distance in the analysis of the spatially varying relationships between house price and several related variables. Furthermore, a Minkowski approach is adopted to select an optimal distance metric for GWR models \citep{lu2016minkowski}.  \cite{lu2018improvements} designed parameter-specific distance metrics instead of ordinary distance to generate spatial weights for observations in local regressions. In addition to the spatial modification of distance metrics, \cite{huang2010stgwr} extends the GWR to the temporal dimension by using space-time distance functions in the process of weights generation.

Emerging studies concentrate on leveraging deep learning techniques to solve the spatial regression problem in a data-driven paradigm. \cite{du2020gnnwr} designed a spatial weighted neural network to generate spatial weights from the distances between samples. Further, the neural network is trained by integrating the spatial weights with the ordinary GWR process to fit the true values. \cite{hagenauer2022gwann} proposed a geographically weighted artificial neural network which applies a geographically weighted error function in the training process, considering the distance between the observation and the location of neurons. By investigating the characteristics of spatial regression models and the graph convolutional network and the similarity between them, \cite{zhu2021srgcnn} developed a spatial regression graph convolutional network to produce reasonable predictions for unobserved locations. However, deep learning methods require massive amounts of high-dimensional spatial data to calibrate the model for better performance. For thousands of data, machine learning techniques with prior knowledge may have an advantage over purely data-driven deep learning models. Additionally, deep neural networks use multiple linear transforms to approximate the fitting targets with a great number of parameters, which can easily lead to overfitting problems. The most critical issue is that neural networks can only complete the task of fitting and prediction, losing the essential capability of explicit quantification for localized observations.

\subsection{Gradient boosting method}
As a type of ensemble learning method, gradient boosting algorithms obtain competitive results in various applications \citep{bentejac2021comparative}. Generally, the gradient boosting method develops an additive model by applying a variety of basic models to learn the current gradient of objective functions to simulate the steepest descent process \citep{friedman2001greedy}. Further, \cite{friedman2002stochastic} proposed a stochastic gradient boosting paradigm that randomly draws a subsample from data instead of using full data set directly to improve the robustness against over-fitting issues. Additionally, numerous studies use a gradient boosting approach to improve base machine learning models, especially tree models. \cite{chen2016xgboost} proposed XGBoost, which applies a decision tree as the basic model. XGBoost computes the second-order approximation of the objective and leverages the regularization term to avoid over-fitting. Moreover, XGBoost designs an exact greedy algorithm and sparsity-aware method to find effectively the optimal split for tree nodes. \cite{ke2017lightgbm} implemented LightGBM, a computationally efficient algorithm of tree-based gradient boosting learning. LightGBM uses the histogram of input data to search for the best split for tree nodes. In addition, LightGBM applies techniques, including gradient-based one-side sampling and exclusive feature bundling, to compromise computational overhead and model performance. \cite{prokhorenkova2018catboost} presented CatBoost, which aims to solve the prediction shift problem in the process of model building. To obtain unbiased gradient estimates, CatBoost trains an individual model for each sample. Thus, CatBoost use ordered boosting to replace the gradient estimate algorithm in the traditional method, which computes the gradient from the data set, excluding the current sample for the training of the base model.

Despite the robustness and efficiency of gradient boosting tree models, they can not explicitly quantify the relationship between independent and dependent variables like generalized linear models in the form of coefficients. In this study, we intend to combine the gradient boosting algorithm and generalized linear model to propose a learning model to generate explicit quantification for spatial variables.

\section{Methodology}
\label{sec:method}
\subsection{Additive linear model for located observations}
In the classic geographically weighted model, an independent linear function is applied to formulate the relationships between dependent and independent variables for each observation $i$ at the specific location:
\begin{equation}
	y_{i}=\beta_{0}\left(u_{i}, v_{i}\right)+\sum_{k=1}^{n} \beta_{k}\left(u_{i}, v_{i}\right) x_{i k}+\varepsilon_{i}
\end{equation}
where $(u_i,v_i)$ denotes the location of $i-$th observation, $\beta_{k}(k=0,\dots,n)$ represents the spatial-varying coefficients of the intercept and independent variables, and the $\varepsilon_{i}$ is the random noise which usually follows a Gaussian distribution hypothesis. The observations in the neighborhood region of location $i$ are collected to estimate the parameters $\hat{\beta}\left(u_{i}, v_{i}\right)$ in a geographically weighted least square approach:
\begin{equation}
	\hat{\beta}\left(u_{i}, v_{i}\right)=\left(X^{\mathrm{T}} W\left(u_{i}, v_{i}\right) X\right)^{-1} X^{\mathrm{T}} W\left(u_{i}, v_{i}\right) y
\end{equation}
where $W(u_i, v_i) \in \mathbb{R}^{N \times N}$ indicates the diagonal spatial weights matrix generated by kernel functions for total $N$ observations. Generally, the off-diagonal elements of $W(u_i, v_i)$ are zero, and the diagonal elements of $W(u_i, v_i)$ denote the geographical weight of $N$ samples for the specific observation $i$. The coefficients of the same independent variable vary across the study area because of respective estimates at each sample location. And such coefficients can be used in further spatial analysis.

Although the GWR coefficients at locations of each observation are estimated in a weighted approach independently, the model performance is evaluated overall with AIC and R$^2$ instead. It may not achieve the global optimal with respective optimization for each local model. Inspired by the ensemble learning paradigm and gradient boosting method, we propose GWRBoost, which uses the localized additive model that leverages the GWR as the base procedure and gradient boosting optimization method for global optimization. For each observation $i$, an additive model is applied to investigate the relationship between variables:
\begin{equation}
	F_i(x) = \sum_{m = 1}^{M} f(x;\beta^m)
	\label{eq:addtive}
\end{equation}
where $f(x;\beta^m)$ is a linear model estimated by the geographically weighted least square algorithm. Thus, the $F(x)$ can be represented explicitly as a uniform linear function:
\begin{equation}
	y_{i} = F_i(x_i) = \sum_{m=1}^{M} \beta_{0}^{m} + \sum_{m=1}^{M} \sum_{k = 1}^{n} \beta_{k}^{m}\left(u_{i}, v_{i}\right) x_{i k} + \varepsilon_{i}
	\label{eq:linear}
\end{equation}

From equation~\ref{eq:linear}, the GWRBoost model is still a linear function which retains the ability to analyze the distribution of parameters and explore the degree of spatial heterogeneity on the basis of global optimization.

\subsection{Geographically weighted gradient boosting}
To estimate the optimal $F(x)$ for observations, the forward stagewise algorithm and gradient boosting method are applied in a greedy manner to meet the requirement of evaluation metrics. Firstly, to be consistent with evaluation metrics, the squared error $L_i$ for the observation $i$ is defined as follows:
\begin{equation}
	L_i = \frac{1}{2} \sum_{j = 1}^{N} (y_j - F_i^M(x_j))^2
	\label{eq:loss}
\end{equation}
where $w_i$ is the $i$-th diagonal element of the spatial weights matrix. The conventional least square algorithm is not applicable for the direct optimization of the comprehensive additive model. However, stage-wise optimization algorithms like the gradient descent can be applied to solve this problem:
\begin{equation}
	\theta_{i} = \theta_{i - 1} - \lambda \frac{\partial L}{\partial \theta_{i - 1}}
\end{equation}
where $\theta_i$ denotes the optimal parameters in the $i$ step and $L$ is the objective error for optimization, and $\lambda$ is the learning rate adjustment factor. For the additive model, the gradient boosting paradigm considers the objective error minimization problem as a complete gradient descent process, taking the $F(x)$ as an optimizable parameter. Thus, the unknown $F^M(x)$ is optimized in a stage-wise approach like general variables in the gradient descent algorithm:
\begin{equation}
	F^m = F^{m - 1} - \lambda \frac{\partial L}{\partial F^{m - 1}}(m = 2,\dots, M)
\end{equation}
where $\lambda$ is the learning rate adjustment factor. In the definition of the local additive model (See Formula~\ref{eq:addtive}),
\begin{equation}
	F^m = F^{m - 1} + f_{\beta^m}(m = 2,\dots, M)
\end{equation}
Thus, to approximate the gradient descent optimization process, each component $f_{\beta^m}$ of the additive model is supposed to learn the current negative gradient value. With square error as the objective function(See equation~\ref{eq:loss}), the gradient of the GWRBoost model is the residual of the last prediction and target values. Therefore, in step $m$, a GWR is trained to fit the gradient value:
\begin{equation}
	f_{\beta^m_i}(x_i) \sim \lambda \frac{\partial L}{\partial F^{m - 1}_i} = \lambda (y_i - F^{m - 1}_i(x_i))
	\label{eq:gradient}
\end{equation}
With the objective function of squared error, the gradient value is the residual of the predictive value.

In summary, the GWRBoost process is given in Algorithm~\ref{algo:boost}. In the first step, we initialize a GWR process for each located sample to generate prediction values for the dependent variable. Then, prediction residuals are computed and new GWR models are trained to fit the residual values continuously until the whole model collects enough base procedure for the final fitting. In practice, an early stopping method can be applied to avoid overfitting problems.
\begin{algorithm}
	\SetAlgoLined

	\caption{Gradient boosting optimization process for localized observations}\label{algo:boost}
	\KwData{Located observations D = $\{(X_1, y_1, u_1, v_1), \dots, (X_N, y_N, u_N, v_N)\}$}
	\KwResult{Linear model set $\mathcal{F^{M}}$ = $\{F^{M}_1, \dots, F^{M}_N\}$}
	\For{$n = 1$ \KwTo $N$}{
	$\hat{\beta}^1_{n} = \mathop{\arg\min}_{\beta_n^1}  \frac{1}{2} \sum_{i = 1}^{N} w_{i} (y_i - f_{\beta_n^1}(x_i))^2$

	$F^1_i = f_{\hat{\beta}_n^1}$
	}
	\For{$m = 2$ \KwTo $M$}{
	\For{$n = 1$ \KwTo $N$}{
	$r_n = \lambda \cdot [y_n - F^{m-1}_n(x_n)]$

	$\hat{\beta}^m_n = \mathop{\arg\min}_{\beta^m_n}  \frac{1}{2} \sum_{i = 1}^{N} w_{i} (r_n - f_{\beta^m_n}(x_i))^2$

	$F^m_{n} = F^{m - 1}_{n} + f_{\hat{\beta}^m_n}$
	}
	}
\end{algorithm}

\subsection{Computation of Akaike information criterion}
The AIC and AICc are the most common metrics to evaluate the fit performance of the GWR model, which are an unbiased estimate of the expected Kullback-Leibler information and a trade-off between goodness of fit and the degree of freedom. In a fitting task, we cannot split the data set into train/test subdivisions because all observations should be analyzed to generate coefficients that quantify the relationship. And it is intractable to measure the degree of overfitting. In AIC/AICc, model complexity is used to represent the possibility of overfitting. A lower value of AIC or AICc indicates the model that has a better fitness performance and a relatively lower parameter complexity. The AIC is defined as follows:
\begin{equation}
	\text{AIC} = - 2 ln(\hat{\mathcal{L}}) + 2 k
\end{equation}
where $\hat{\mathcal{L}}$ indicates the maximum value of log likelihood function of the model and $k$ is the number of effective parameters, which implies the model complexity. Generally, most linear regressions tend to apply Gaussian to model the uncertainty of estimate results. In terms of the degrees of freedom, the generalized linear model applies the trace of hat matrix $tr(\mathcal{H})$ to represent $k$. The hat matrix $\mathcal{H}$ is defined as the transform matrix that maps dependent variable $y$ to the model prediction output $\hat{y}$:
\begin{equation}
	\hat{y} = \mathcal{H} y
    \label{eq:hatmat}
\end{equation}
In the GWR, which is composed of individual linear regression estimates from locations of all observations respectively, each row $r_i$ of ultimate hat matrix $\mathcal{H}$ is retrieved from the corresponding row of the hat matrix $S_i$ for the observation $i$ \citep{fotheringham2003gwr}.

However, it is not appropriate to apply the computation of the number of effective parameters for GWR to the GWRBoost model directly because it is not a generalized linear model. Here we formulate the approach to compute the degree of freedom in the GWRBoost. First, the original hat matrix $\mathcal{H}$ is constant in each iteration because it is estimated in the same spatial weights and observations $X$. Therefore, the prediction target in the $m$ iteration can be derived from the initial iteration:
\begin{equation}
	\begin{aligned}
		y_{m} & = y_{m - 1} - \hat{y}_{m - 1}                        \\
		      & = \lambda (I - \mathcal{H}) y_{m - 1}                \\
		      & = \left[\lambda (I - \mathcal{H})\right]^{m - 1} y_1
	\end{aligned}
\end{equation}
The total fit target can be summarized as:
\begin{equation}
	\begin{aligned}
		\hat{y} & = \sum_{m=1}^{M} \hat{y}_m                                                                     \\
		        & = \sum_{m=1}^{M} \mathcal{H} y_m                                                               \\
		        & = \mathcal{H} \sum_{m=1}^{M} (I - \mathcal{H}) ^ {m - 1} y_1                                   \\
		        & = \left\{\mathcal{H} \sum_{m=1}^{M} \left[\lambda (I - \mathcal{H})\right] ^ {m - 1}\right\} y
	\end{aligned}
	\label{eq:boost_hat}
\end{equation}
By the definition of hat matrix (See Equation~\ref{eq:hatmat}), we can derive the hat matrix of $M$ total iterations in GWRBoost:
\begin{equation}
    \mathcal{H}^{M} = \mathcal{H} \sum_{m=1}^{M} \left[\lambda (I - \mathcal{H})\right] ^ {m - 1}
\end{equation}
Furthermore, the AIC can be calculated with the Gaussian log likelihood function and the given degree of freedom.
\begin{equation}
	\text{AIC} = - 2 ln(\hat{\mathcal{L}}) + 2 tr(\mathcal{H}^{M})
\end{equation}
Besides, the AICc value of GWRBoost can be computed conveniently in the performance evaluation, unlike other complex nonlinear machine learning algorithms. And precise comparative analysis can be conducted to show the theoretical and practical values of the proposed model.

\section{Experimental results}
\label{sec:expr}
\subsection{Evaluation metrics}
For evaluation, we use six metrics to investigate the performance of the model, which include:
\begin{itemize}
	\item Root mean square error (RMSE):\begin{equation}
		      \mathrm{RMSE}_j=\sqrt{\frac{1}{n} \sum_{i=1}^n\left(\beta_j\left(u_i, v_i\right)-\hat{\beta}_j\left(u_i, v_i\right)\right)^2}
	      \end{equation}
	      where $\beta_{j}\left(u_{i}, v_{i}\right)$ is the $j$-th coefficient and $\hat{\beta}_{j}\left(u_{i}, v_{i}\right)$ is the coefficient estimates at location $(u_i, v_i)$.
	\item Residual sum of squares (RSS):\begin{equation}
		      \text{RSS} = \sum_{i=1}^{n} (y_i - F^M(x_i))^2
		      \label{eq:rss}
	      \end{equation}
	\item R$^2$:\begin{equation}
		      \text{R}^2 = 1 - \frac{\sum_{i=1}^n (y_i - F^M(x_i))^2}{\sum_{i=1}^n (y_i - \bar{y})}
	      \end{equation}
	      where $\bar{y}$ is the mean value of response variable $y_i(i = 1,\dots,n)$ of all observations.
	\item Adjusted R$^2$:\begin{equation}
		      \text{Adjusted R}^2 = 1 - \frac{(n - 1)(1 - R^2)}{n - k - 1}
	      \end{equation}
	      where $k$ is the degree of freedom.
	\item AIC: \begin{equation}
		      \text{AIC} = - 2 ln(\hat{\mathcal{L}}) + 2 k
		      \label{eq:aic}
	      \end{equation}
	      where $\hat{\mathcal{L}}$ is the maximum estimate of log likelihood and $k$ is the degree of freedom.
	\item AICc:\begin{equation}
		      \text{AICc} = - 2 ln(\hat{\mathcal{L}}) + 2 k + \frac{2k(k+1)}{n - k - 1}
	      \end{equation}
	      where $n$ is the number of observations and $k$ is the degree of freedom.
	\item Moran's I:\begin{equation}
                I=\frac{N \sum_{i=1}^N \sum_{j=1}^N w_{i j}\left(x_i-\bar{x}\right)\left(x_j-\bar{x}\right)}{W \sum_{i=1}^N\left(x_i-\bar{x}\right)^2}
        \end{equation}
        where $w_{i j}$ denotes the spatial weight of sample $i$ and $j$.
\end{itemize}
The RMSE indicates the accuracy of local coefficients estimation accuracy \citep{fotheringham2017mgwr}. A lower RMSE implies that a more accurate parameter set is estimated by the evaluated model. The RSS and R\textsuperscript{2} aim to evaluate the goodness of fit of the dependent variable $y$ for the model on a global data scale. A lower RSS and a higher R\textsuperscript{2} mean a better fit and prediction. The Adjusted R\textsuperscript{2} and AIC combines the criteria of goodness of fit and the model complexity to make a comprehensive evaluation. A higher adjusted R\textsuperscript{2} and a lower AIC implies a more effective model with a lower model complexity, which is desirable because of both better fitting performance and less risk of overfitting. Further, AICc corrects AIC to adapt to smaller data sets. The Moran'I indicator implies the degree of spatial autocorrelation of distributed values. In this study, we use Moran's I, which ranges in [-1, 1], for the evaluation of predicted residuals. The value below 0 implies a negative dispersion pattern, and above 0 implies a positive and agglomeration pattern. The spatial distribution of residuals from an effective fitting process is supposed to show no significant dispersion or agglomeration patterns, as Moran's I is around $\frac{-1}{N - 1}$ where $N$ denotes the number of observations.


The R$^2$ and Adjusted R$^2$ metric both contain the term $\sum_{i=1}^n (y_i - F^M(x_i))^2$ which denotes the minimum of prediction square error \textit{RSS}. Furthermore, the estimated log likelihood term $\hat{\mathcal{L}}$ in the AIC and AICc metric can be converted into square error in the general Gaussian hypothesis for the random noise term $\varepsilon$:
\begin{equation}
	p(\mathbf{y} \mid \mathbf{x}, \mathbf{\beta}, \sigma)=\mathcal{N}\left(y \mid F^{M}(\mathbf{x}, \mathbf{\beta}), \sigma^{-1}\right)
\end{equation}
where $\sigma$ is the inverse variance of the random term. Thus, the log likelihood value $\hat{\mathcal{L}}$ can be estimate as:
\begin{equation}
	\begin{aligned}
		\ln p(\mathbf{y} \mid \mathbf{\beta}, \sigma) & =\sum_{i=1}^n \ln \mathcal{N}\left(y_i \mid F_{\beta}\left(\mathbf{x}_i\right), \sigma^{-1}\right)         \\
		                                              & =\frac{n}{2} \ln \sigma-\frac{n}{2} \ln (2 \pi)-\sigma \cdot \frac{1}{2} \sum_{i=1}^n (y_i - F^{M}(x_i))^2 \\
		                                              & =\frac{n}{2} \ln \sigma-\frac{n}{2} \ln (2 \pi)-\sigma \cdot \text{RSS}
	\end{aligned}
	\label{eq:loglikelihood}
\end{equation}

Therefore, common criteria for spatial regressions aim to evaluate the sum of square error (See Formula \ref{eq:rss}, \ref{eq:aic} and \ref{eq:loglikelihood}). The square loss objective function is supposed to be consistent with the evaluation metrics and treats all localized observations equally without additional weights to achieve better performance. Thus, GWRBoost with squared loss as an objective function can be proven to be theoretically effective.

\subsection{Validation on simulation data}
\begin{figure}[htbp]
	\centering
	\includegraphics[width=1\linewidth]{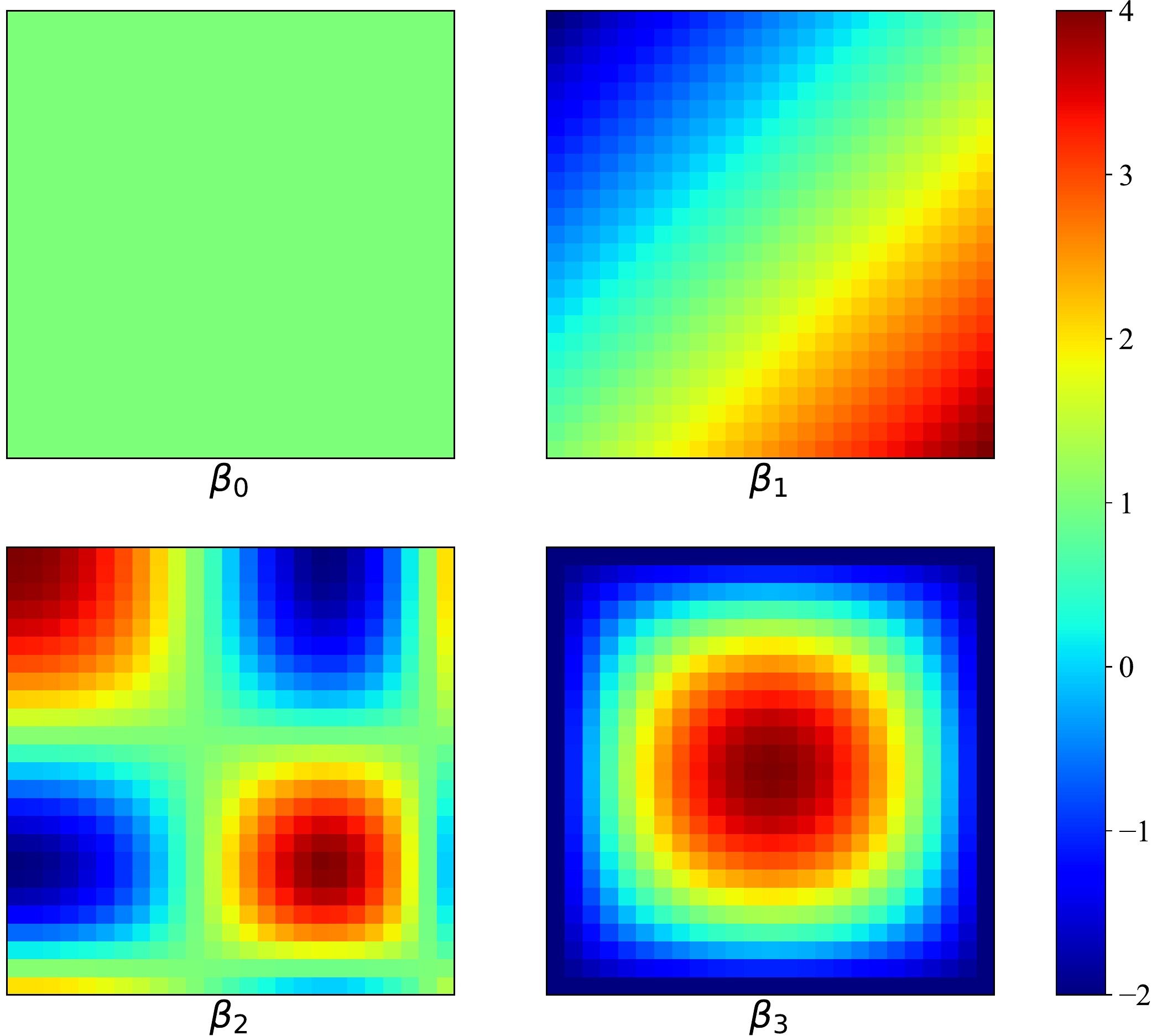}
	\caption{The distribution of simulation coefficients}
	\label{fig:coef}
\end{figure}

In this section, we apply simulation data to verify the performance of GWRBoost. A square grid dataset of $25 \times 25$ cells is generated, where all coefficients are determined by the fixed location function. Like \cite{lu2018improvements, fotheringham2017mgwr}, we adopt complex simulation design for coefficients with different degrees of heterogeneity as follows:
\begin{itemize}
	\item Stationary:\begin{equation}
		      \beta_0(u_i, v_i) = 2
	      \end{equation}
	\item Low heterogeneity:\begin{equation}
		      \beta_1(u_i, v_i) = \frac{1}{8}\left(u_i + v_i\right) - 2
	      \end{equation}
	\item Medium heterogeneity:\begin{equation}
		      \beta_2(u_i, v_i) = 3 \cdot \operatorname{cos}(\pi\operatorname{e}^{u_i / 25}) \operatorname{sin}(\pi\operatorname{e}^{v_i / 25}) + 1
	      \end{equation}
	\item High heterogeneity:\begin{equation}
		      \beta_3(u_i, v_i) = \frac{1}{216}\left[36-\left(6-\frac{u}{2}\right)^{2}\right]\left[36-\left(6-\frac{v}{2}\right)^{2}\right] - 2
	      \end{equation}
\end{itemize}
where $(u_i, v_i)$ is the horizontal and vertical location of observation $i$, which range in $[1, 25]$. The values of generated coefficients range in $[-2, 4]$. The distribution of coefficients is visualized in Figure~\ref{fig:coef}. Then, the corresponding formula to generate the response variable $y$ is
\begin{equation}
	y =\beta_0 + \beta_1 x_1 + \beta_2 x_2 + \beta_3 x_3 + \varepsilon
\end{equation}
where $x_i (i = 1,2,3)$ is randomly sampled from uniform distribution $\mathcal{N}(0, 1)$ and the noise term is sampled from the normal distribution $\mathcal{N}(0, 0.5^{2})$. 100 individual data sets are generated independently to examine the robustness and stability of the proposed model and the consistency of the experimental results.

In this experiment, we compare the performances of ordinary least square (OLS), GWR and GWRBoost for all 100 simulation data sets. The GWR adopts optimal bandwidth with the AICc score in each data set. For the GWRBoost, we apply a larger bandwidth than the original optimal one of  the initial GWR by multiplying an adjustment factor to introduce more local observations to mitigate the potential overfitting issue. The bandwidth of GWRBoost is fixed in each optimization step. In addition, the early stopping technique is applied by measuring the change in R\textsuperscript{2}/AICc metrics. If the score is worse in the new learning step, the model will output the result of the previous step as the final optimal estimation. 

In general, the comparative results show that our proposed model can reduce the RMSE by 18.3\% in parameter estimation accuracy and AICc by 67.3\% in the goodness of fit than classic GWR, which indicates the GWRBoost is a more effective method to evaluate the spatially varying relationships between localized variables.

\begin{figure}[htbp]
	\centering
	\includegraphics[width=1\linewidth]{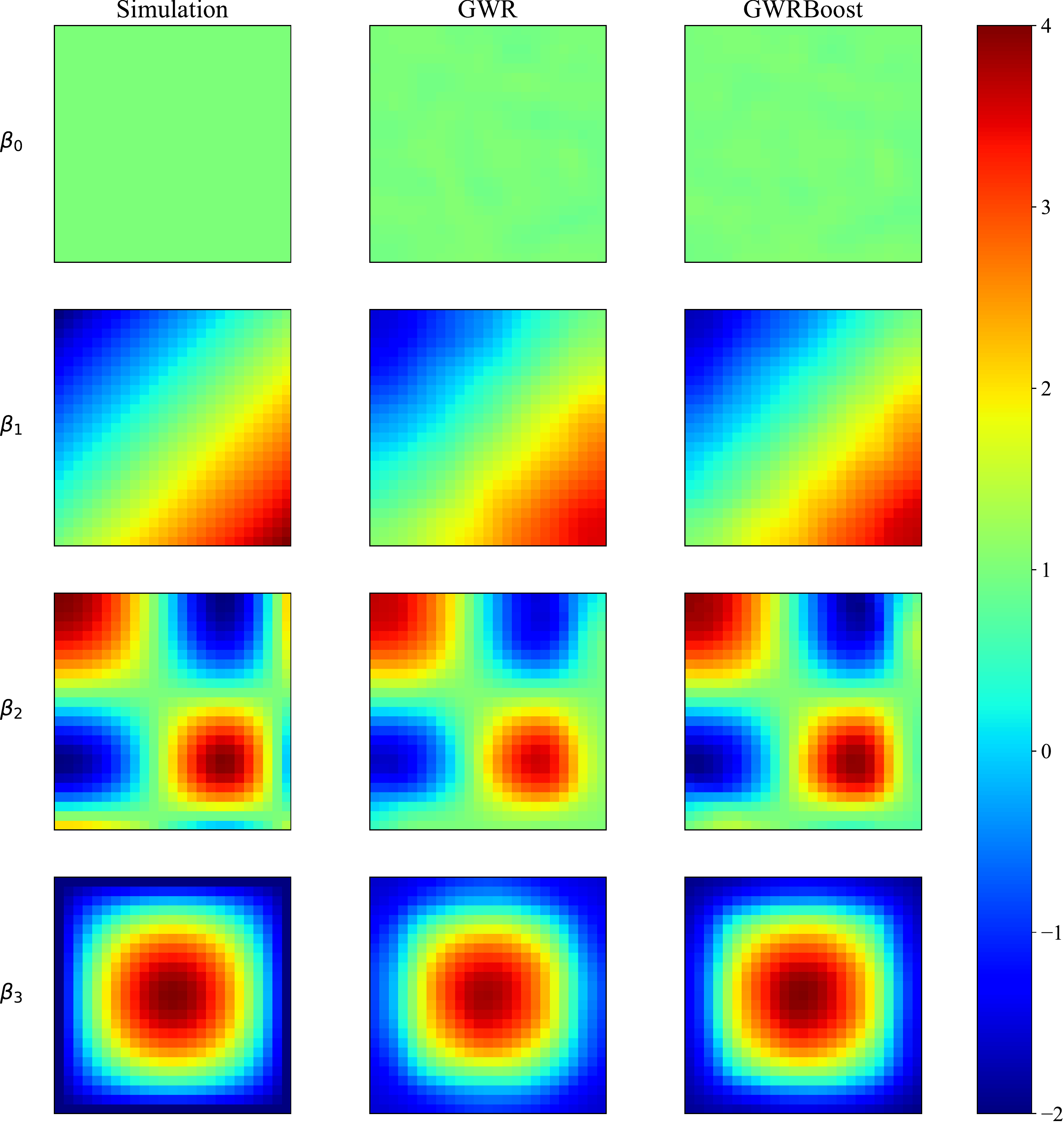}
	\caption{Comparative results of parameter distribution}
	\label{fig:param}
\end{figure}

Figure~\ref{fig:param} shows the comparison between ground truth and  parameter estimates generated by GWR and GWRBoost. The results show that both GWR and GWRBoost achieve a good estimate for spatially varying coefficients. However, for $\beta_1$, $\beta_2$, $\beta_3$, GWR outputs an inferior performance than GWRBoost in the estimates of maximum and minimum values, which indicates that GWR maintains a relatively lower model complexity and learns a more smooth fitting result with a smaller range of values. Insufficient parameters are used in GWR to estimate in the simulation data that has high spatial heterogeneity. Whereas GWRBoost has a better estimate of such maximum/minimum values, which implies that GWRBoost holds the appropriate capacity of parameters to capture the internal heterogeneous characteristics of localized observations.

\begin{figure}[htbp]
	\centering
	\includegraphics[width=1\linewidth]{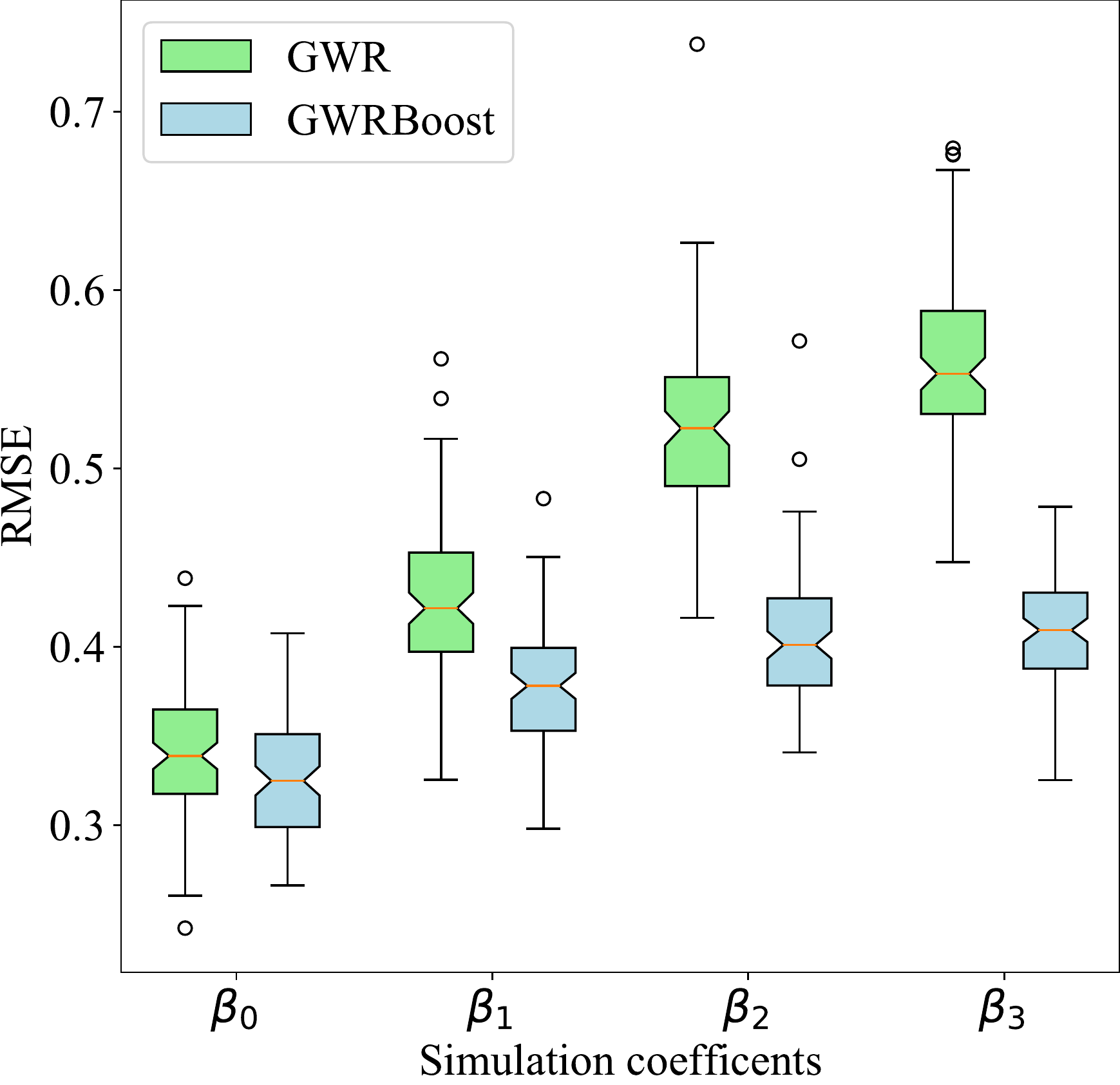}
	\caption{RMSE of coefficients estimates}
	\label{fig:rmse}
\end{figure}

Figure~\ref{fig:rmse} shows the RMSE value of coefficients estimates for the GWR and GWRBoost, where each box describes the statistical distribution of estimates error from 100 simulations. The estimates for coefficients are far more important than the goodness of fit because these coefficients from linear models represent explainable quantification of the relationship. The comparative result indicates that GWRBoost can generate more precise estimates for localized relationships than GWR without potential issues of overfitting or underfitting. Additionally, compared to GWR, GWRBoost effectively reduces the estimation error of parameters with higher heterogeneity and remain a relatively low output variance. 

\begin{figure}[htbp]
	\centering
	\includegraphics[width=1\linewidth]{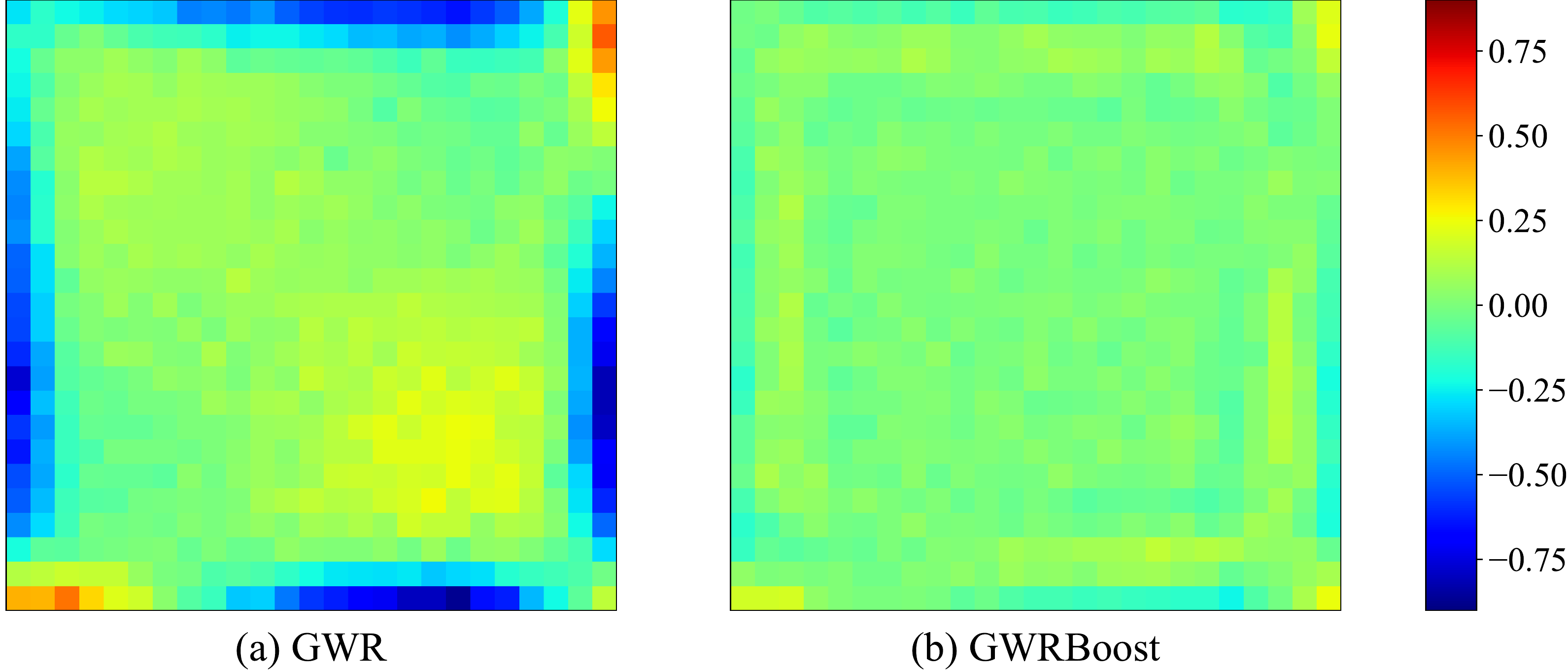}
	\caption{Spatial distribution of residuals in simulation experiment}
	\label{fig:moran}
\end{figure}

The spatial distribution of predicted residuals is shown in Figure~\ref{fig:moran}. It is noticeable that GWRBoost outputs lower and more uniformly distributed residuals than GWR. In the small edge region of the simulation data, GWR produces higher residuals that clearly have a strong spatial autocorrelation of agglomeration, while in the centre part two larger areas of weaker agglomeration can be seen. On the contrary, residuals from GWRBoost show no significant spatial pattern of agglomeration or dispersion, which indicates that it produces an effective fitting result. T accurate degree is evaluated by the Moran's I later.

\begin{table*}[ht]
	\centering
	\caption{Comparative performance of simulation data}
	\begin{tabular*}{\hsize}{@{\extracolsep{\fill}}lrrr}
		\toprule
		Model                         & OLS                   & GWR                    & \textbf{GWRBoost}               \\ \midrule
		RSS                           & 1639.063 $\pm$ 72.52  & 83.900 $\pm$ 5.049     & \textbf{36.797 $\pm$ 2.601}     \\
		AIC                           & 2385.642 $\pm$ 27.65  & 773.374 $\pm$ 36.050   & \textbf{225.512 $\pm$ 42.061}   \\
		AICc                          & 2385.739 $\pm$ 27.65  & 839.926 $\pm$ 35.383   & \textbf{274.817 $\pm$ 41.207}   \\
		R\textsuperscript{2}          & 0.072 $\pm$ 0.02      & 0.952 $\pm$ 0.003      & \textbf{0.979 $\pm$ 0.002}      \\
		Adjusted R\textsuperscript{2} & 0.066 $\pm$ 0.02      & 0.940 $\pm$ 0.004      & \textbf{0.975 $\pm$ 0.002}      \\ 
		Moran's I                     & 0.712 $\pm$ 0.017 & 0.242 $\pm$ 0.026 & \textbf{-0.064 $\pm$ 0.017}\\
		\bottomrule
	\end{tabular*}
	\label{tab:performance}
\end{table*}

The result of the comparison is shown in Table~\ref{tab:performance}. All results in the table are presented as the mean value $\pm$, the standard deviation value of 100 generated simulations. It is obvious that OLS produces relatively inferior performance while GWR detects the spatial heterogeneity in the impacts of independent variables on dependent variables and outperforms the OLS significantly. However, better scores of GWRBoost imply that it produces a more accurate fitting of parameters and prediction output for spatialized data observations. In terms of the degree of spatial autocorrelation for residuals, OLS and GWR show a strong spatial pattern of agglomeration, while the residuals of GWRBoost maintain a random distribution, which supports the idea that it captures more heterogeneous characteristics of spatially-varying relationships than others. Furthermore, the lower variance of RSS and R\textsuperscript{2} indicates that GWRBoost can learn the spatial characteristics of relationships regardless of the random input. However, the large variance of AIC and AICc means that the complexity of the parameters is different for each simulation experiment because each model is precisely adapted to the specific data set.

In summary of the simulation experiment, GWRBoost can both estimate the coefficients more effectively and provide a better fit for the dependent variable. In addition, it reduces the spatial agglomeration of predicted residuals and maintains a relatively robust output than other algorithms.

\subsection{Empirical case study}
In this section, we aim to use the NYC education data set provided by GeoDa Lab (\url{https://geodacenter.github.io/data-and-lab//NYC-Census-2000}) to show the practical value of our proposed method. The NYC education data set collected educated information on citizens in New York City in 2000, which uses block as the basic unit. The data set contains 2,216 blocks of different sizes, covering an area of 783.170 km$^2$ (8429.975 ft$^2$), and 56 independent variables derived from census data, including ethnicity, gender and age share.

\begin{figure}[htbp]
	\centering
	\includegraphics[width=1\linewidth]{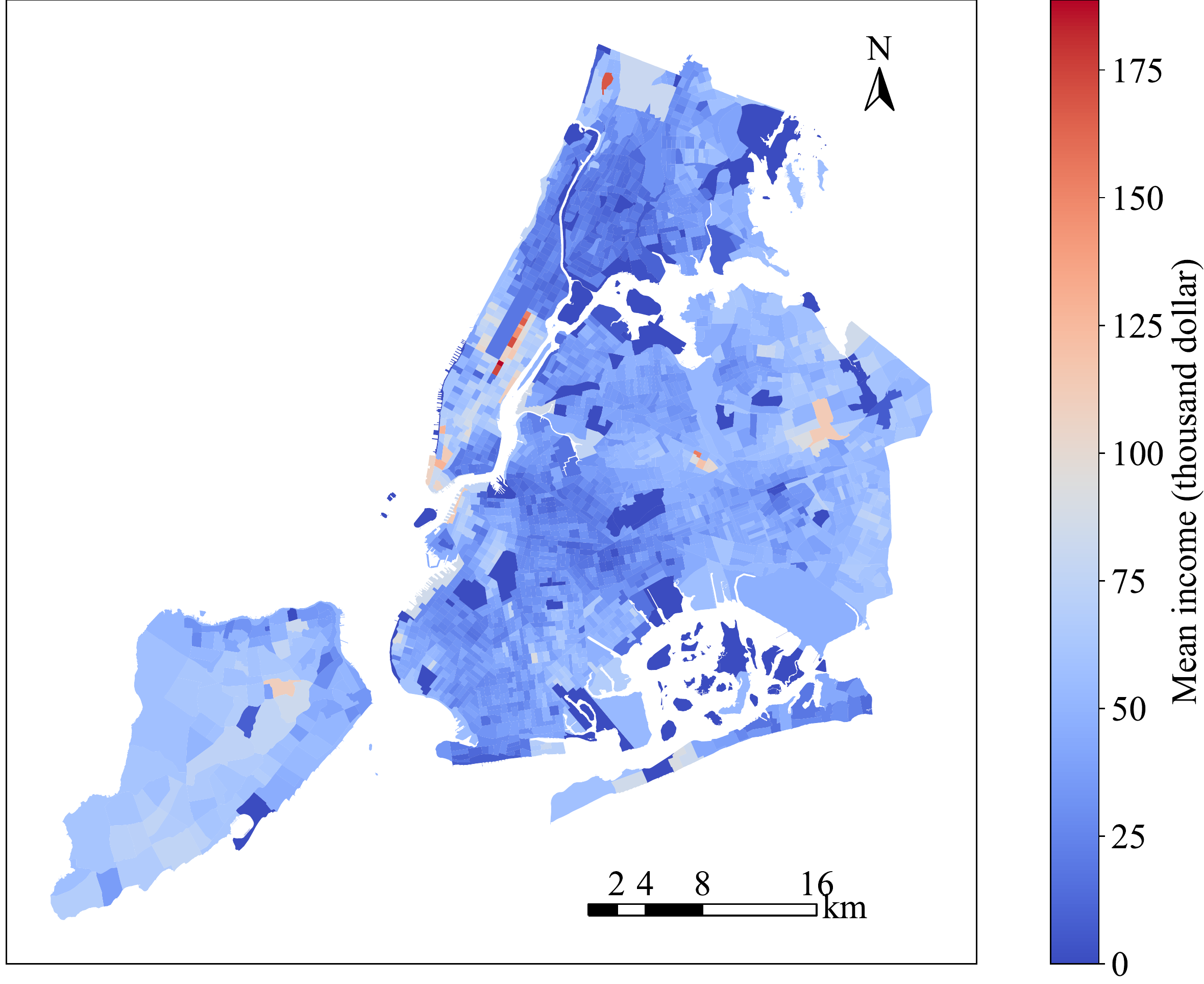}
	\caption{Spatial distribution of mean income in NYC}
	\label{fig:nyc}
\end{figure}

In the empirical case study, to demonstrate the ability to handle massive and complex data of the proposed model, a more difficult scenario than the simulation is designed. We aim to explain the degree of educational status of the residents on the income per capita. Figure~\ref{fig:nyc} shows the spatial distribution of individual mean income in the study area of NYC. It is clear that the average income in most regions is below 100 thousand dollars while that of some areas can reach 125 thousand dollars or higher. People on Fifth avenue at East of Central Park in Manhattan have the most revenue of 188.7 thousand dollars in NYC and substantially higher than surrounding areas. Similar to Fifth avenue, places near the Henry Ittleson centre in Fieldston and the neighbourhood of the West Slide Tennis Club in Forest Hills are inhabited by high-income citizens. Significant spatial variation also causes difficulties for the subsequent regression analysis. In addition, a certain degree of spatial agglomeration can be qualitatively identified in different parts. 

 Some numerical indicators in the topic of education are selected as the independent variables for the quantitative analysis of effects on the mean income in NYC as Table~\ref{tab:var} shows. We mainly apply measurements of educated individuals in different age groups, such as the percentage of the population who dropped out of high school or obtained at least a bachelor's degree, to explain the effects on mean income.

\begin{table}[htbp]
\centering
\caption{Selected indicators}
\begin{tabularx}{\hsize}{@{\extracolsep{\fill}}lX}
\toprule
\textbf{Variable}        & \textbf{Explanation}\\ \midrule
\textbf{Dependent}   &                                                                                 \\
mean\_inc            & Mean income                                                                     \\ \addlinespace
\textbf{Independent} &                                                                                 \\
sub18                & Population under 18 years (count)                                         \\ \addlinespace
PER\_PRV\_SC         & Percentage of all students enrolled in private school                           \\ \addlinespace
YOUTH\_DROP          & Percentage of population age 16-19 that has dropped out of high school          \\ \addlinespace
HS\_DROP             & Percentage of population age over 25 that dropped out of high school            \\ \addlinespace
COL\_DEGREE          & Percentage of population age over 25 that obtained at least a bachelor’s degree \\ \addlinespace
SCHOOL\_CT           & Number of schools (count) \\     
\bottomrule
\end{tabularx}
\label{tab:var}
\end{table}

We first perform a z-score transformation on the dependent variable and independent variables. Then the bi-square function is applied to generate the spatial weight. The same metrics in the simulation experiment are used to evaluate the fitting performance of OLS, GWR, and GWRBoost. After searching with the AICc metric, the optimal bandwidth for GWR is 100. The results are shown in Table~\ref{tab:case}. It is clear that GWR shows a substantially better performance than OLS, which implies that spatial heterogeneity is significant in the study area. GWRBoost outperforms other algorithms in all metrics, especially the AIC/AICc. The model complexity increases because of more base linear models are added, but the overall score is better due to the substantial improvement of fitting performance. Furthermore, Moran's I for the residuals from GWRBoost is more close to the random distribution that has a theoretical value of -0.0004, which indicates the simple linear regression is inadequate and a more advanced model is required to learn the complex spatial varying relationship between variables.

\begin{table}[htbp]
	\centering
	\caption{Comparative performance of NYC education data set}
	\begin{tabular*}{\hsize}{@{\extracolsep{\fill}}lrrr}
		\toprule
		Model                         & OLS                   & GWR                    & \textbf{GWRBoost}               \\ \midrule
		RSS                           & 982.206  & 388.626 & \textbf{261.478}     \\
		AIC                           & 4499.669  & 3168.118 & \textbf{2289.994}   \\
		AICc                          & 4499.720 & 3315.637 & \textbf{2437.513}   \\
		R\textsuperscript{2}          & 0.557 & 0.825 & \textbf{0.882}      \\
		Adjusted R\textsuperscript{2} & 0.556 & 0.790 & \textbf{0.858}      \\ 
		Moran's I                     & 0.333 & 0.066 & \textbf{-0.027}\\
		\bottomrule
	\end{tabular*}
	\label{tab:case}
\end{table}

\begin{figure}[htbp]
	\centering
	\includegraphics[width=1\linewidth]{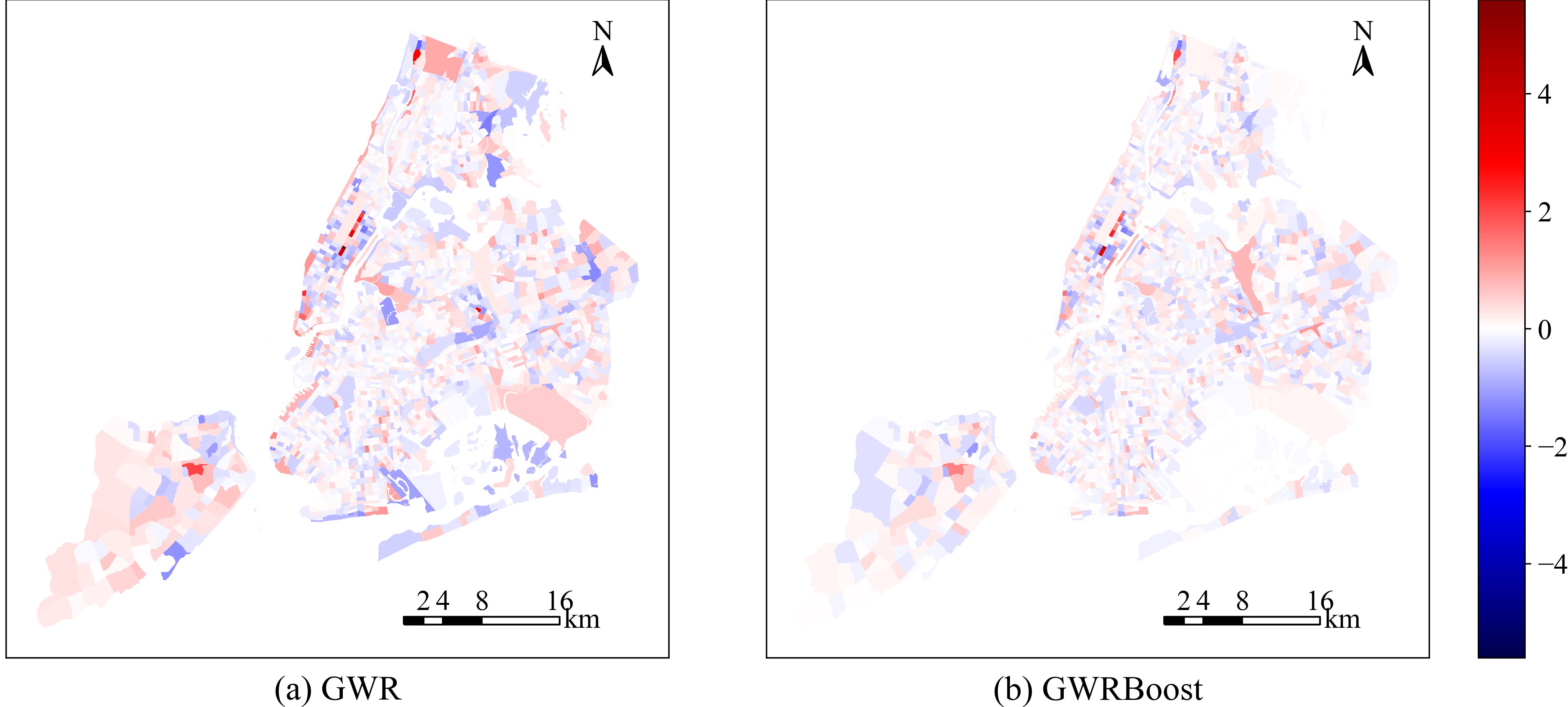}
	\caption{Spatial distribution of model residuals}
	\label{fig:nyc_resid}
\end{figure}

Figure~\ref{fig:nyc_resid} demonstrates the spatial distribution of output residuals from GWR and GWRBoost. Compare to GWR, the residuals from GWRBoost are overall smaller. In some parts of the northeast and southeast, the residuals are nearly 0. The mean values of residuals from GWR and GWRBoost are close to 0 but the maximum deviates more from 0 than the minimum because the estimates for Fifth Avenue are much lower than the ground truth.

Table~\ref{tab:summary} shows the statistical summary of estimates for spatially varying coefficients. Generally, the percentage of all students enrolled in private school (PER\_PRV\_SC), the percentage of people who own at least a bachelor's degree (COL\_DEGREE), and the number of schools (SCHOOL\_CT) have a positive impact on the mean income while the effects of the population under 18 (sub18), the percentage of the population age 16-19 that has dropped out of high school (YOUTH\_DROP), and the percentage of population age over 25 that dropped out of high school (HS\_DROP) are negative. In addition, COL DEGREE plays the most significant role in the positive relationship with average earnings, at the level of 0.521. It is obvious that employers with higher education are more likely to earn higher wages. The count of schools in the neighbourhood has a slight influence on the mean income, which is only one percent of COL\_DEGREE, because the supply of educational services does not directly indicate the educated status of citizens. The contribution of YOUTH\_DROP is also insignificant that it only affects the mean income when these children grow up and begin to look for a job.

\begin{table}[htbp]
\centering
\caption{Statistical summary of GWRBoost estimates}
\begin{tabular*}{\hsize}{@{\extracolsep{\fill}}lrrrrr}
\toprule
Variable & Mean & Min & Max & Std \\ \midrule
Intercept & -0.126 & -2.723 & 1.685 & 0.619 \\
sub18 & -0.118 & -1.967 & 1.750 & 0.335 \\
PER\_PRV\_SC & 0.169 & -1.824 & 3.258 & 0.465 \\
YOUTH\_DROP & -0.019 & -1.743 & 2.579 & 0.260 \\
HS\_DROP & -0.083 & -2.828 & 3.511 & 0.587 \\
COL\_DEGREE & 0.521 & -2.228 & 3.468 & 0.783 \\
SCHOOL\_CT & 0.005 & -0.952 & 0.905 & 0.213 \\
\bottomrule
\end{tabular*}
\label{tab:summary}
\end{table}

\begin{figure*}[ht]
	\centering
	\subfloat[sub18]{
	    \includegraphics[width=0.33\linewidth]{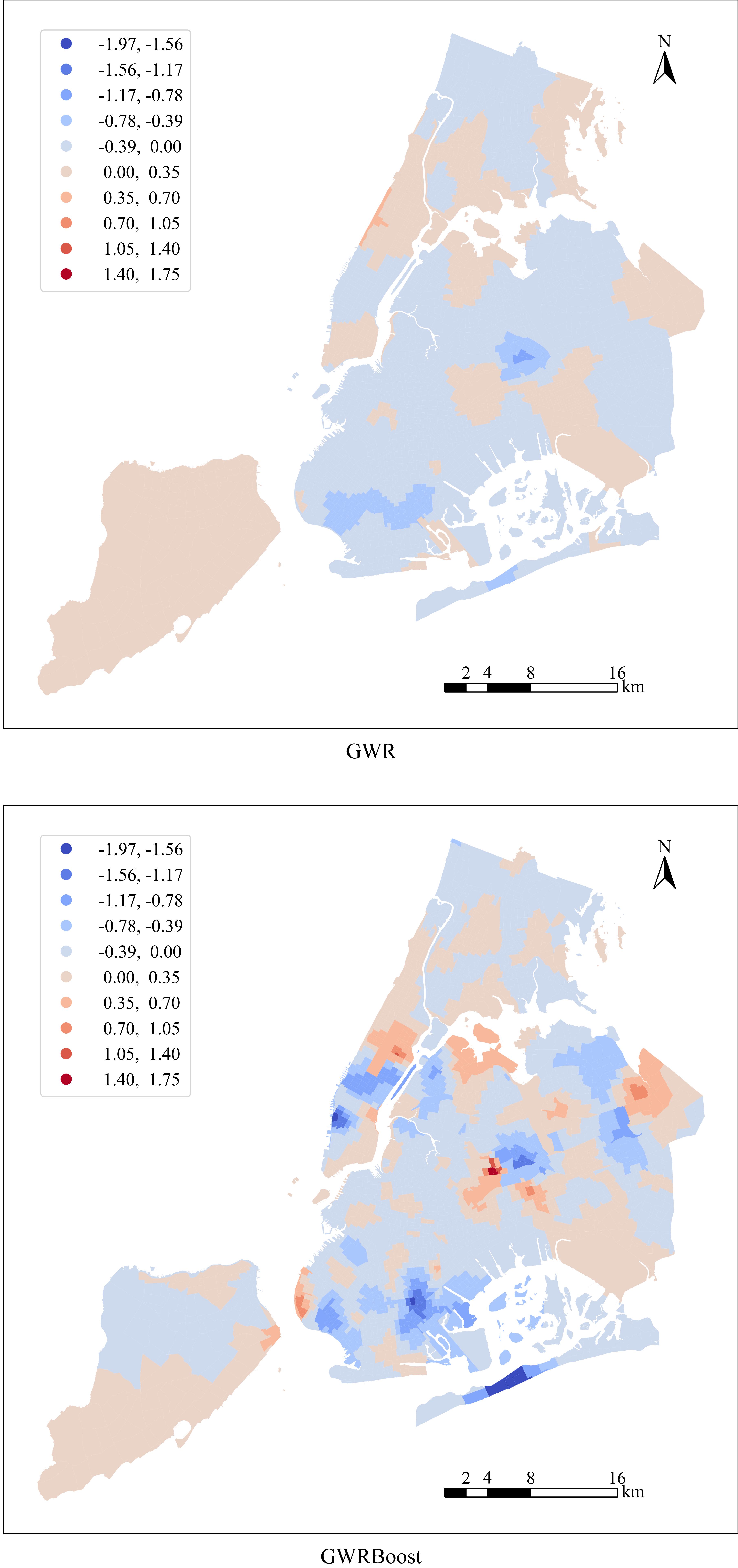}
	}
    \subfloat[HS\_DROP]{
	    \includegraphics[width=0.33\linewidth]{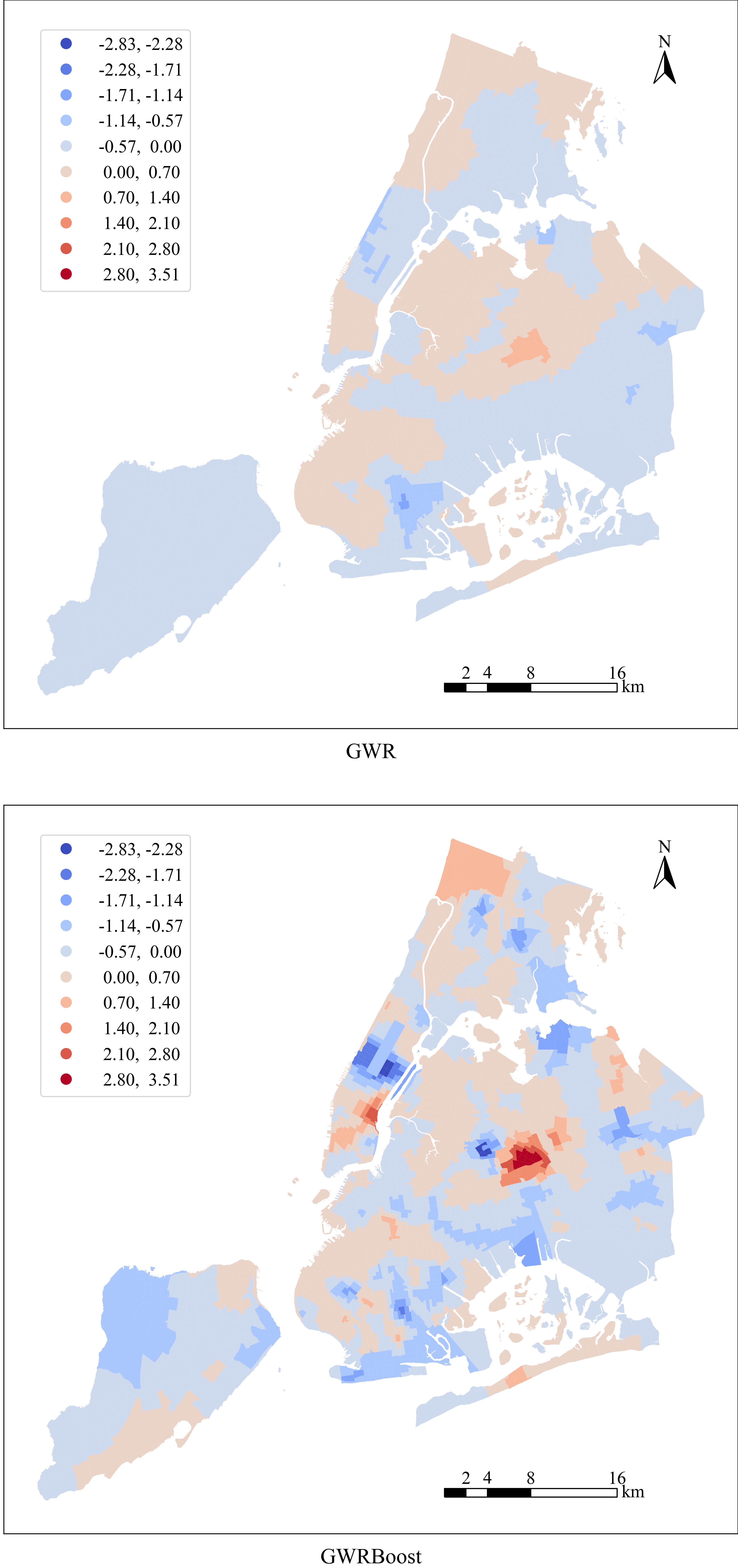}
	}
    \subfloat[COL\_DEGREE]{
	    \includegraphics[width=0.33\linewidth]{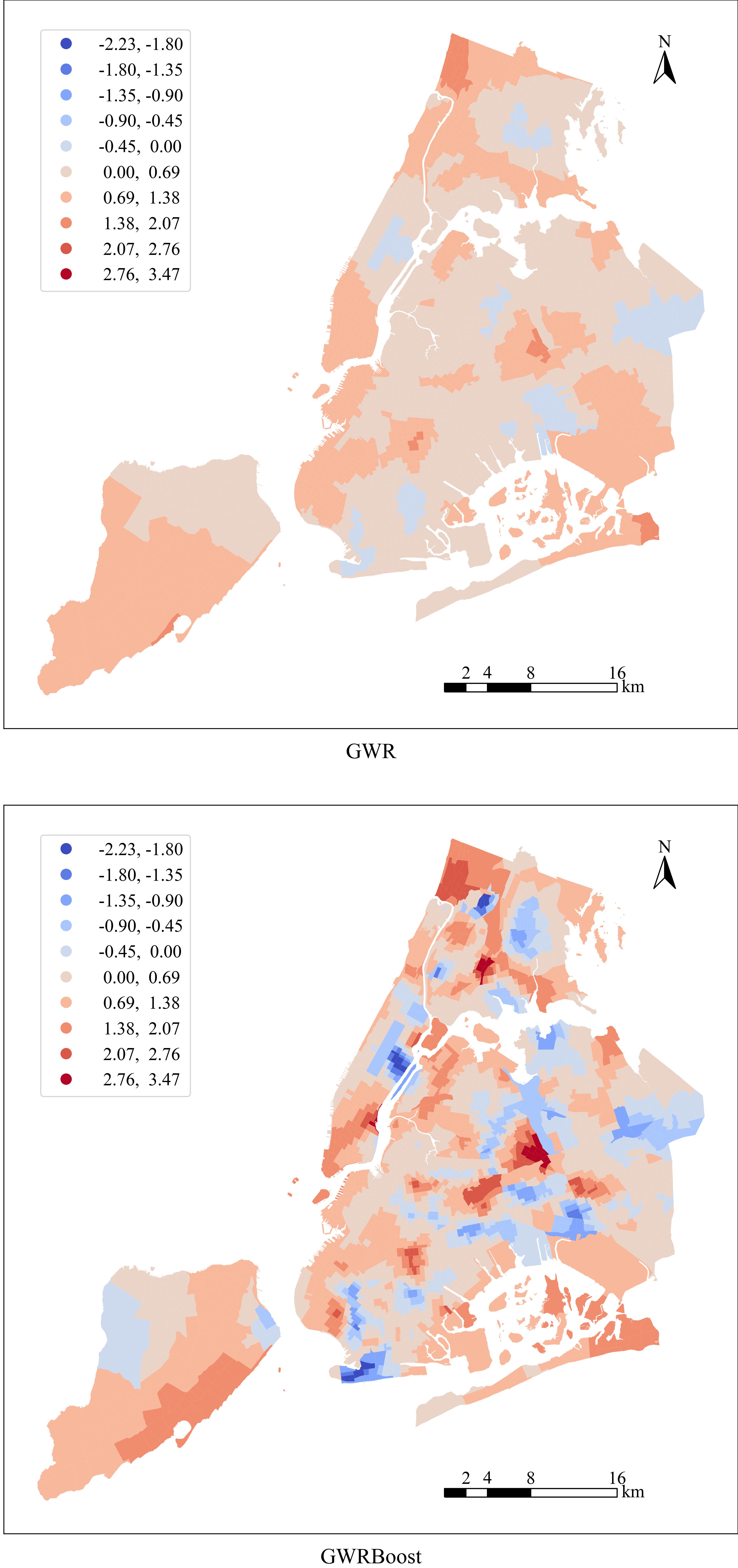}
	}
	\caption{Spatial variation of selected coefficient estimates}
	\label{fig:nyc_coef}
\end{figure*}

The comparison of spatial variation of local coefficients estimates for selected indicators, including sub18, HS\_DROP and COL\_DEGREE, is further demonstrated in Figure~\ref{fig:nyc_coef}. Coefficients are primarily divided into positive and negative parts, which are further split into 5 equal categories respectively. Overall, the coefficients estimated by GWRBoost are similar to that of GWR but vary in a larger range.

It is noticeable that the impacts of sub18 are mainly negative. Unemployed children are not paid, thus reducing per capita income. However, wealthier parents in some places are eager to raise more offspring. Such a phenomenon is easily smoothed in global regression analysis but will be clearly detected in local models. Even after being average with many children, the income per capita in affluent areas is still higher than in their neighbourhood. GWRBoost explains this relationship by producing larger parameters than GWR in some of the top-income areas, including Fifth Avenue in Manhattan and the neighbourhood of the West Slide Tennis Club in Forest Hills.

Intuitively, HS\_DROP is supposed to have a negative effect on the average income of individuals. However, findings of both GWR and GWRBoost show that it has a two-way effect. Especially in the bustling central area of Manhattan, it is difficult for high school dropouts to find suitable jobs with high salaries there, which leads to dramatic negative coefficient estimates in GWRBoost. However, some labour-intensive jobs with substantial income do not require a bachelor's degree or more. Workers who have dropped out can also earn high wages. In the neighbourhood of Forest Hills, Kew Gardens Hills and Flushing, HS\_DROP poses a positive impact on incomes. One possible reason is that a great number of immigrants with low education can also earn high incomes through business or catering services.

In terms of COL\_DEGREE, GWR and GWRBoost reach a similar result that it has a significant positive influence almost across the whole area, which indicates that well-educated citizens can earn more salaries. However, regions of Park Avenue in Manhattan maintain negative relationships between the mean income and both COL\_DEGREE and HS\_DROP. Compared to their neighbourhood, the local mean income is excessively high with strong spatial heterogeneity. Neither COL\_DEGREE nor HS\_DROP may not be the primary contributing factor. The analysis is required to be further meticulously scrutinized due to the issue of omitted variable bias.

\begin{figure*}[ht]
	\centering
	\subfloat[bandwidth adjustment 0.8]{
	    \includegraphics[width=0.33\linewidth]{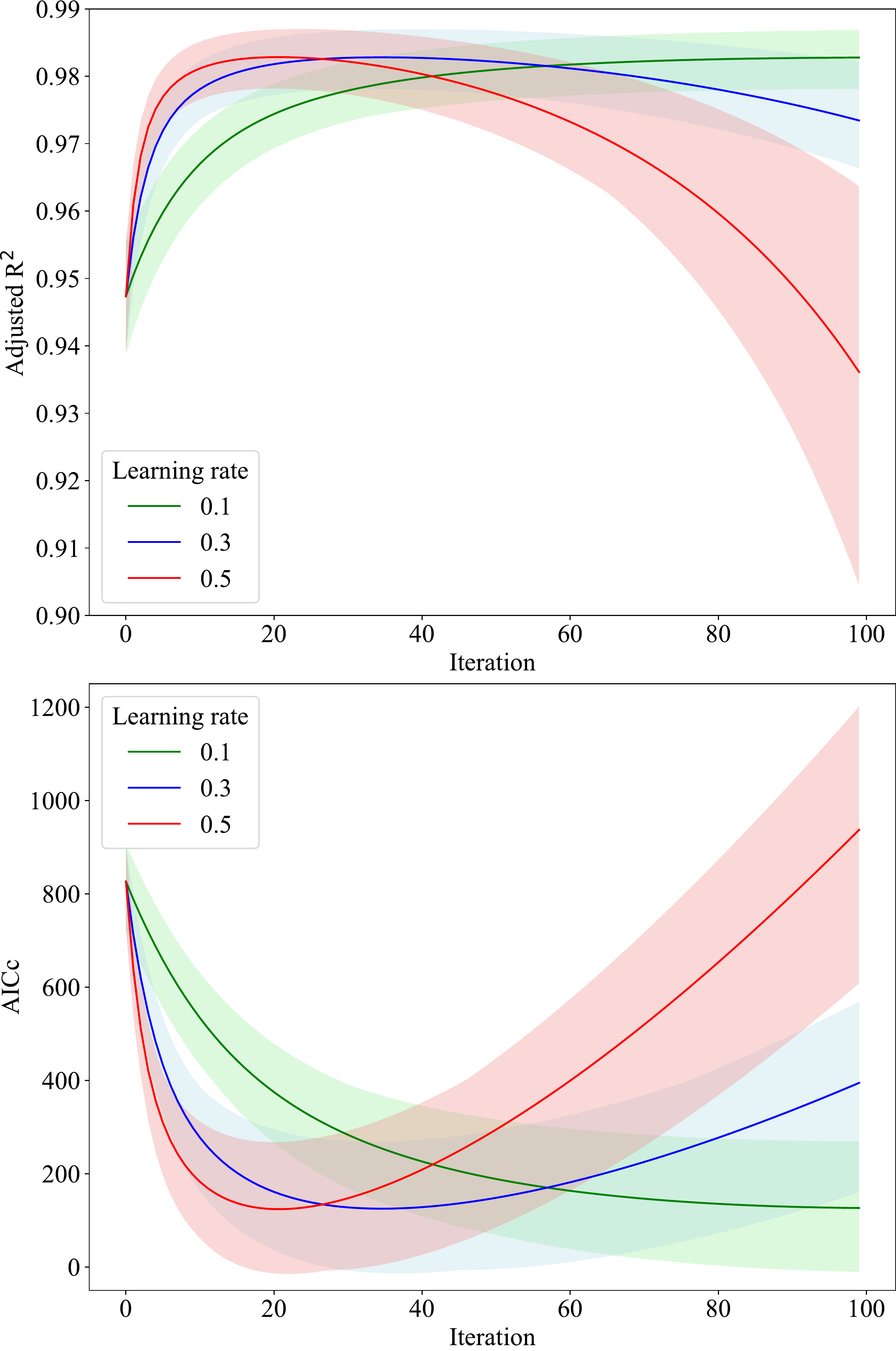}
	}
	\subfloat[bandwidth adjustment 1]{
	    \includegraphics[width=0.33\linewidth]{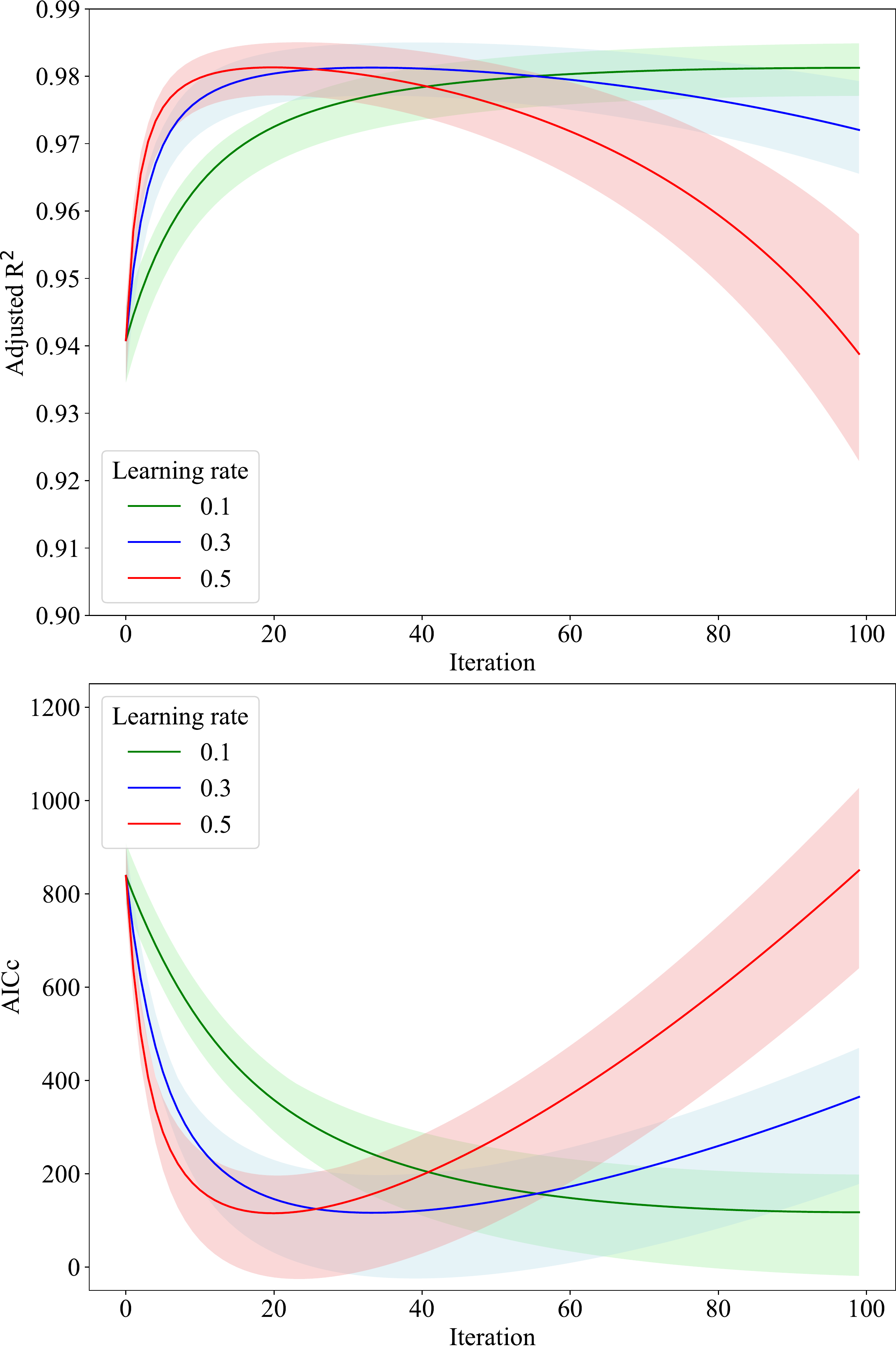}
	}
	\subfloat[bandwidth adjustment 1.2]{
	    \includegraphics[width=0.33\linewidth]{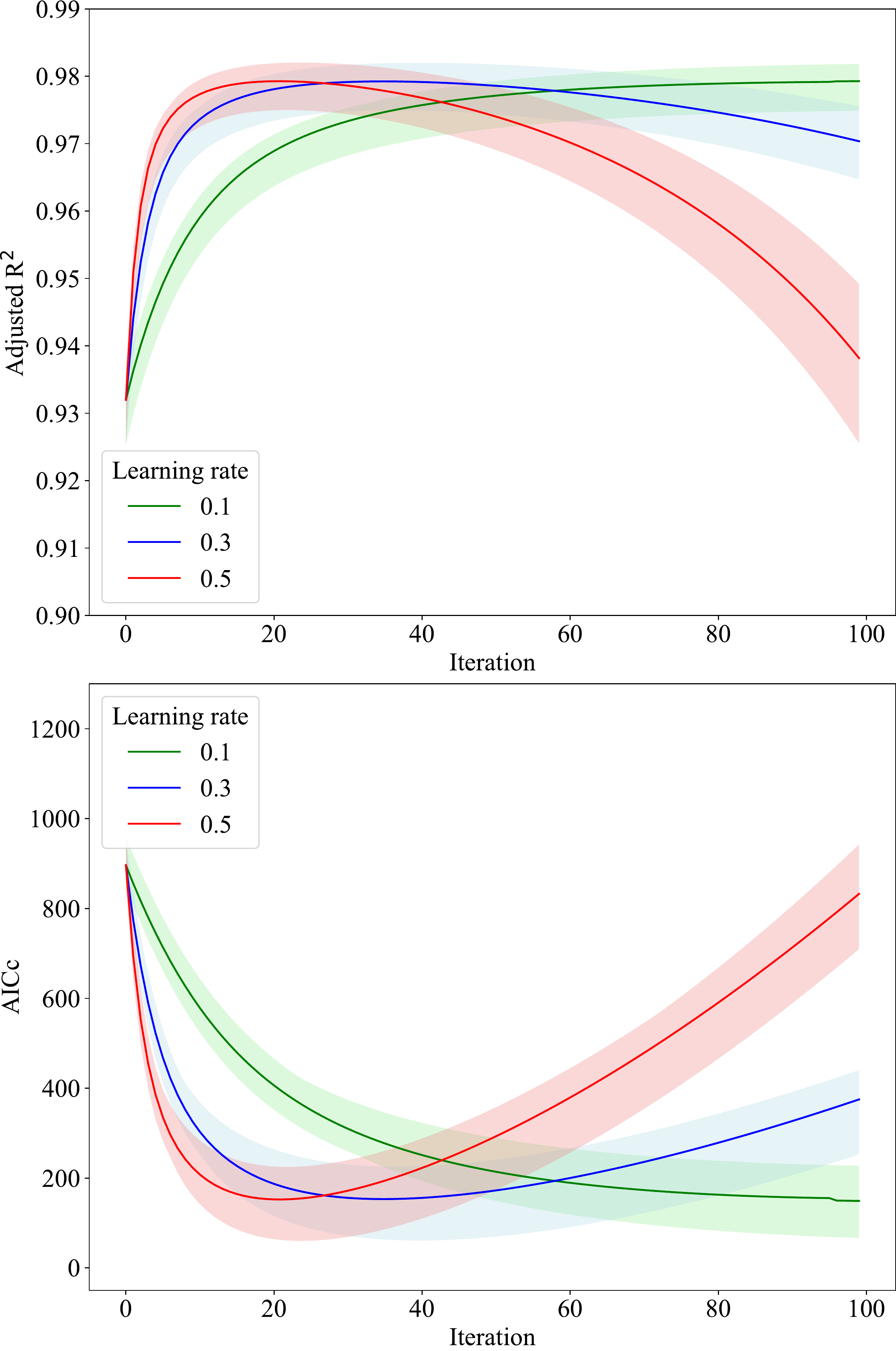}
	}
	\caption{Regression results with different iteration numbers, learning rate and bandwidth adjustment}
	\label{fig:overfitting}
\end{figure*}

In general, GWRBoost generates wider ranges of parameters than GWR because it has a higher degree of model complexity to produce a better fit for dependent variables. Furthermore, the coefficients vary more significantly on small scales to capture the spatial heterogeneity effectively, which is similar to regression results with small bandwidths in GWR. However, GWRBoost with larger bandwidths can avoid issues of ill condition or matrix singularity in solving the equation.

\section{Discussion}
\label{sec:discuss}
\subsection{Sensitivity analysis of hypterparameters}
In the general ensemble learning process, the overfitting problem will be more significant with the increase of base learners. The number of base learners and the learning rate per learner holds are crucial to the performance of output results. To investigate the sensitivity of hyperparameters, including the number of learners, learning rate and the bandwidth adjustment factor, we conducted a variety of additional experiments on 100 randomly generated data sets. The results are shown in Figure~\ref{fig:overfitting}. 

It is clear that a certain number of learners can reach optimal results while more or fewer of them may lead to inferior performance. The boosting model first achieves the optimum and then becomes worse. Thus, it is necessary to set the early stopping technique to monitor and output the optimal learning result.

With regard to the learning rate, we can see that it can determine when the model starts to be overfitting. In general, a larger learning rate contributes to achieving worse results rapidly. The smaller learning rate can reach a peak value that is slightly better than larger ones but it takes much longer. More sophisticated and adaptive learning schedulers, such as the warm-up scheduler or the cosine annealing scheduler, can be applied to improve the learning process. 

In terms of the problem of bandwidth selection, the optimal value has not yet been identified. The result shows that smaller bandwidth only allows fewer neighbouring samples to participate in the regression, which lead to the significant overfitting problem and unstable output with the increase of learners and model complexity. Whereas a larger bandwidth with more samples can mitigate the issue. However, too large a bandwidth deprives the localized regression of its original purpose. Therefore, an efficient search algorithm is further required in the enhancement of GWRBoost. Additionally, a multi scale version of GWRBoost can be further developed to investigate the importance of scale in spatial regression models.

\subsection{Effectiveness of the gradient boosting optimization}
Ensemble learning is an effective tool in data science. It gathers some trained models from weak machine learning algorithms to compose a strong one. Gradient boosting learning uses simple learners as the stage-wise optimization method, especially in an intractable and non-parameter regression problem. Since the least square achieves the optimal solution of linear regression, it is useless to use a complex gradient boosting method that only approaches the optimum. However, in a spatial regression problem, we can consider the collection of regressors for each localized observation as a whole non-parameter model and apply the gradient boosting optimization to find the optimum effectively. 

The first reason is that the composition of multiple basic learners increases the model complexity and capacity, which enables the model to learn more patterns and characteristics from massive and heterogeneous data than classic linear regression. Furthermore, the whole model leverages the same metric used in the evaluation instead of a biased one. In GWR, each linear model at the location of a specific observation obtains the optimal parameters by a weighted objective function. From the perspective of the whole model, it is difficult to ensure that all local optimum can constitute the global optimal parameters. Nevertheless, GWRBoost searches the optimum for each observation with the same unweighted least square objective function and a weighted gradient boosting optimization. Although the global optimum may not be obtained as in the case of least squares, it can be effectively approached by the gradient descent algorithm. In addition, few observations fed into the model by small bandwidth filtering can cause overfitting easily. In GWRBoost, each model at a certain location applies residuals in its neighbouring regions to predict the centre residual value generated by the previous stage. From a particular observation, its neighbouring samples convey information about their own neighbours implicitly through residual values to it. Even more distant observations are still able to transmit the information to the centre through the gradual delivery of residuals. Therefore, each model can capture the global characteristics hierarchically, which actually enhances the performance of GWRBoost.

However, it is widely acknowledged that ensemble learning algorithms share the common issue of intensive computational overhead, especially for boosting models where numerous base learners cannot be trained in parallel. With a large amount of observations needed to fit a model respectively, GWRBoost consumes enormous computational overhead. GWRBoost also suffers from the computation of AIC. Equation~\ref{eq:boost_hat} implies that the hat matrix generated by each stage is collected to compute the final degree of freedom, which indicates that it needs extra memory and computing resources. Thereby a more efficient computation method is required to be further studied. 

\section{Conclusion}
\label{sec:conclusion}
In this study, we propose GWRBoost, a geographically weighted gradient boosting model to quantify spatially-varying relationships between localized variables explicitly. We increase the model complexity by applying the localized additive model and gradient boosting optimization method to alleviate underfitting issues caused by linear models. In addition, GWRBoost retains the capability of explainable relationship quantification by aggregating parameter estimates from all base linear functions. Both the simulation experiment and empirical case study show the efficient performance and practical value of GWRBoost. However, several issues proposed imply that further studies are needed to improve the method. For example, we can investigate the adaptive optimal bandwidth in different stages and the reduction of computational overhead.

\bibliography{mybibfile}

\end{document}